\documentclass[conference]{IEEEtran}
\IEEEoverridecommandlockouts
\usepackage{cite}
\usepackage{makecell}
\usepackage{amsmath,amssymb,amsfonts}
\usepackage{algorithmic}
\usepackage{graphicx}
\usepackage{textcomp}
\usepackage{float} 
\usepackage{xcolor}
\usepackage{graphicx,makecell}
\usepackage{float}
\usepackage{subcaption}
\usepackage{tikz}
\usetikzlibrary{arrows.meta,positioning,fit,shadows.blur}
\def\BibTeX{{\rm B\kern-.05em{\sc i\kern-.025em b}\kern-.08em
    T\kern-.1667em\lower.7ex\hbox{E}\kern-.125emX}}
\begin{document}

\title{Kantian--Utilitarian XAI: Meta-Explained}

\author{\IEEEauthorblockN{1\textsuperscript{st} Zahra Atf}
\IEEEauthorblockA{\textit{Faculty of Business and Information Technology} \\
\textit{Ontario Tech University}\\
Oshawa, Canada \\
ORCID: 0000-0003-0642-4341}
\and
\IEEEauthorblockN{2\textsuperscript{nd} Peter R. Lewis}
\IEEEauthorblockA{\textit{Faculty of Business and Information Technology} \\
\textit{Ontario Tech University}\\
Oshawa, Canada \\
ORCID: 0000-0003-4271-8611}
}

\maketitle

\begin{abstract}
We present a gamified explainable AI (XAI) system for ethically aware consumer decision‐making in the coffee domain. Each session comprises six rounds with three options per round. Two symbolic engines provide real-time reasons: a Kantian module flags rule violations (e.g., child labor, deforestation risk without shade certification, opaque supply chains, unsafe decaf), and a utilitarian module scores options via multi-criteria aggregation over normalized attributes (price, carbon, water, transparency, farmer income share, taste/freshness, packaging, convenience). A meta-explainer with a regret bound ($\approx 0.2$) highlights Kantian–utilitarian (mis)alignment and switches to a deontically clean near-parity option when welfare loss is small. We release a structured configuration (attribute schema, certification map, weights, rule set), a policy trace for auditability, and an interactive UI. 
\end{abstract}

\begin{IEEEkeywords}
Explainable AI (XAI), Gamification, Kantian and Utilitarian Logic, Consumer Decision-Making.
\end{IEEEkeywords}

\section{Introduction}
Everyday coffee choices hide moral trade-offs behind simple price tags. A ``cheap'' pod can carry a high carbon and water footprint, opaque labor practices, or risky decaf processing. We build a gamified XAI system that makes these trade-offs explicit in real time: a Kantian module flags deontic violations, a utilitarian module aggregates welfare over ethically salient attributes, and a meta-explainer surfaces alignment/conflict while allowing users to tune value weights.
We compare four explanation conditions---none, Kantian, utilitarian, and combined+meta---across six coffee rounds (three options per round). Attributes are min--max normalized within each round. The coffee attribute schema includes: price (CAD/cup), carbon (gCO$_2$e/cup), water (L/cup), supply-chain transparency (0--1), farmer income share (\%), deforestation risk (0--1) with \texttt{shade\_cert} (bool), child labor risk (0--1), packaging recyclability (0/1) and type (categorical), taste score (0--100), freshness (days since roast), brew time (minutes; convenience proxy), decaf process (\texttt{none, water, co2, solvent\_safe, solvent\_risky}), and vegan certification (bool). Kantian rules encode thresholds/requirements (e.g., R1 child labor, R2 deforestation without shade, R3 low transparency, R4 low farmer income share, R5 risky decaf, R6 irresponsible packaging); the utilitarian engine uses weighted MCDA over normalized features with signs matching ethical directionality (e.g., negative for price/carbon/water, positive for transparency/farmer income/taste).
In summary, relative to a price-only baseline, utilitarian explanations tend to boost welfare while leaving more deontic violations; Kantian explanations eliminate violations at some welfare cost; and the combined+meta condition reduces deontic risk with a modest welfare trade-off and resolves a meaningful share of Kantian-utilitarian conflicts (e.g., 37.5\% in our pilot).
The literature supports this design: human-style explanations (causal, contrastive, comparative) enhance comprehension and acceptance \cite{b1}\cite{b2}; user studies define outcome measures for calibration \cite{b3}\cite{b4}; and moral identity shapes judgments and prosocial choices \cite{b5}\cite{b6}. Gamification shows efficacy for interactive decision-making; consequentialist scoring uses social preferences for culturally sensitive welfare assessment; and deontic methods encode rule constraints to prevent norm violations \cite{b7}\cite{b8}\cite{b9}.Together, these strands motivate our head-to-head test of Kantian, utilitarian, and combined meta-explanations on trust, comprehension, and decision quality in a gamified consumer setting, with value--explanation alignment as a moderator and trust as a mediator.

\section{Research Methodology}
\begin{figure}[H] 
\centering
\begin{subfigure}[t]{0.48\linewidth}
  \centering
  \includegraphics[width=\linewidth]{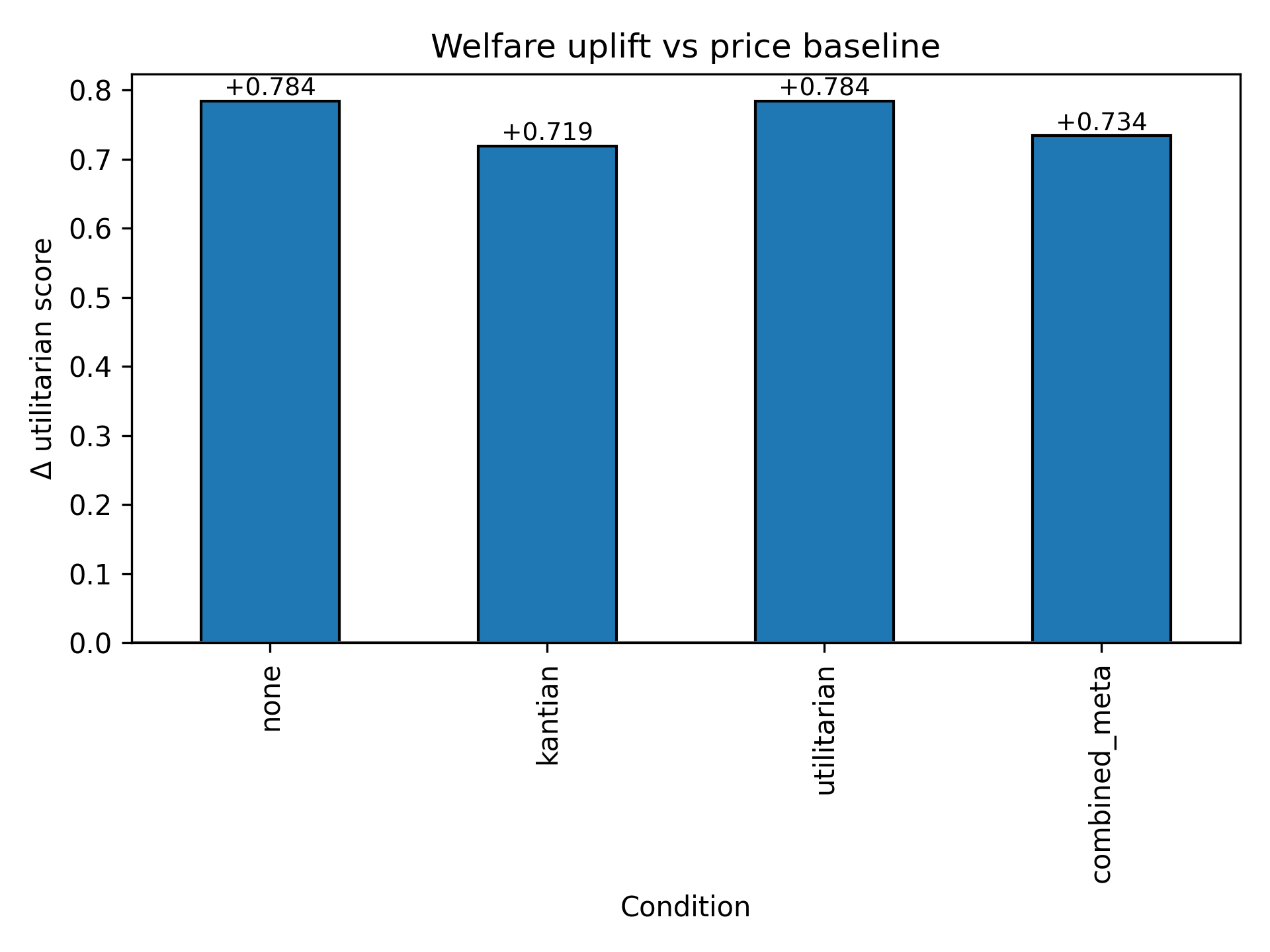}
  \caption{Welfare uplift vs.\ price baseline}\label{subfig:welfare}
\end{subfigure}\hfill
\begin{subfigure}[t]{0.48\linewidth}
  \centering
  \includegraphics[width=\linewidth]{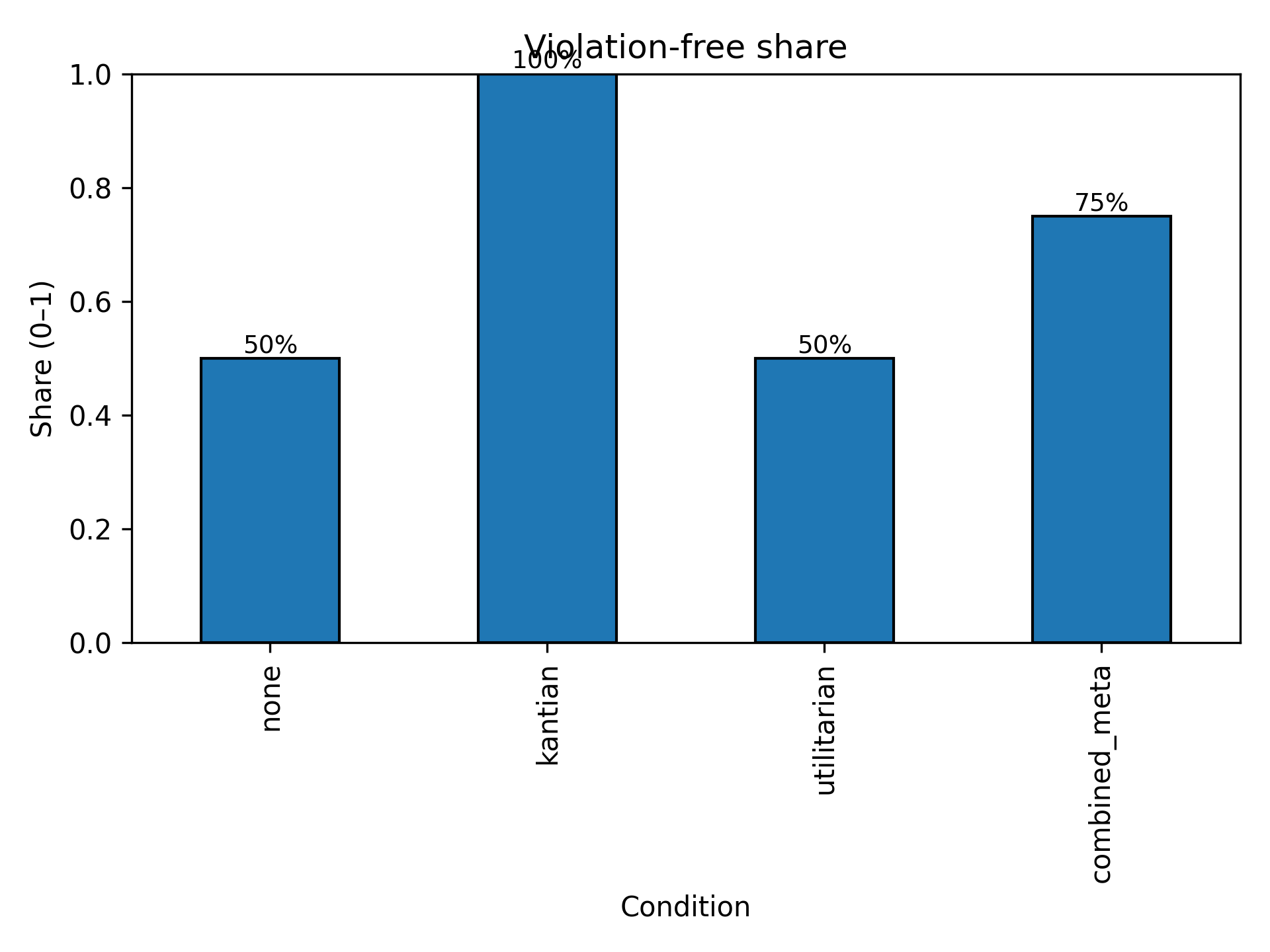}
  \caption{Violation-free share}\label{subfig:violfree}
\end{subfigure}
\caption{Condition-level effects on (a) welfare uplift and (b) violation-free choices across six synthetic coffee rounds. The combined + meta condition preserves near-utilitarian welfare while improving deontic compliance.}
\label{fig:cond_plots}
\end{figure}

We study how explanation type shapes ethically-aware consumer choices in a gamified XAI setting. Participants are operationalized as simulated agents. Agents are randomized to one of four conditions: \textbf{no explanation}, \textbf{Kantian-only}, \textbf{utilitarian-only}, or \textbf{combined + meta-explainer}. Outcomes are compared at the condition level (Table~\ref{tab:cond_summary}; Fig.~\ref{fig:cond_plots}).

\begin{table}[H] 
\caption{Condition-level outcomes on synthetic coffee scenarios. 
$\Delta$ welfare = utilitarian uplift vs.\ price baseline; 
``violation-free'' = share with zero violations; 
conflict resolved = meta-explainer replacements.} \label{tab:cond_summary}
\centering
\setlength{\tabcolsep}{4pt}
\renewcommand{\arraystretch}{1.1}
\resizebox{\linewidth}{!}{%
\begin{tabular}{|p{2.4cm}|c|c|c|c|}
\hline
\textbf{Condition} &
\makecell{\textbf{$\Delta$ welfare}\\\textbf{vs. price}} &
\makecell{\textbf{Violation-}\\\textbf{free (\%)}} &
\makecell{\textbf{Mean Kantian}\\\textbf{severity}} &
\makecell{\textbf{Conflict}\\\textbf{resolved (\%)}} \\
\hline
No explanation   & +0.784 & 50  & 0.875 & --- \\
\hline
Kantian-only     & +0.719 & 100 & 0.000 & --- \\
\hline
Utilitarian-only & +0.784 & 50  & 0.875 & --- \\
\hline
Combined + Meta  & +0.702 & 75  & 0.500 & 25  \\
\hline
\end{tabular}%
}
\end{table}

We run 6 rounds of budgeted coffee shopping. In each round, the agent chooses one product among 3 options drawn from a pool of synthetic coffee scenarios. Four conditions are evaluated: (i) \textit{no explanation}, which picks the option that maximizes the agent’s personalized utility; (ii) \textit{Kantian-only}, which selects a deontically clean option when available (ties resolved by utility) and otherwise minimizes rule-violation severity; (iii) \textit{utilitarian-only}, which maximizes the utilitarian score; and (iv) \textit{combined + meta}, which applies a regret-bounded switch—if the top-utility choice violates Kantian rules but a violation-free option exists within a small utility margin of the per-scenario maximum, the system recommends the clean option; otherwise it keeps the top-utility option.

\noindent\textbf{Stimuli.} Six coffee rounds; each round presents three options constructed from a configurable schema (\texttt{attribute\_schema.json}) and certification map (\texttt{cert\_map.yml}). Per-round min–max normalization ensures within-context comparability. \noindent\textbf{Engines.} \emph{Kantian}: thresholded deontic rules (R1–R6) flag and score violations. \emph{Utilitarian}: a weighted linear aggregate over normalized attributes with signs aligned to ethical direction (e.g., negative: price, carbon, water; positive: transparency, farmer income share, taste).\noindent\textbf{Meta-explainer.} Detects Kantian–utilitarian conflict and applies regret-bounded switching: if the top-utility option violates Kantian rules but a violation-free option exists within the regret margin, the system recommends the clean near-parity option; otherwise it keeps the utility-best. Default regret bound is $0.2$.\noindent\textbf{UI \& logging.} Streamlit front-end with per-round \emph{Why/Details} panels; user picks are logged to \texttt{outputs/play\_log.csv}. The pipeline exports auditable traces (\texttt{policy\_trace\_text.csv}), condition summaries (\texttt{condition\_summary.csv}), and figures. \noindent\textbf{Outcomes.} Welfare uplift (vs.\ price-only), violation-free share, mean Kantian severity. \noindent\textbf{Protocol \& reporting.} Condition means averaged across scenarios under default weights (alt-weights for sensitivity); Table~\ref{tab:cond_summary} (means) and Fig.~\ref{fig:cond_plots} (bar plots). Conflict-resolution rate = fraction where the meta layer replaces a violating utilitarian-best with a violation-free near-parity option. An anonymized, seed-fixed artifact enables exact reproducibility.

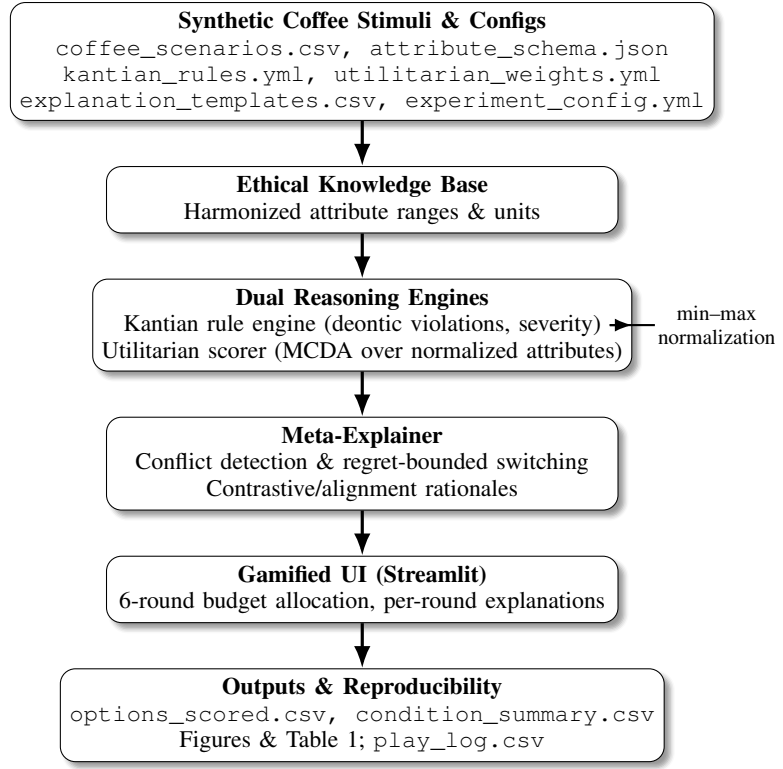
\begin{figure}[t]
\centering
\begin{tikzpicture}[
  font=\small,
  node distance=6mm,
  block/.style = {draw, rounded corners=2mm, align=center, inner sep=3.5pt,
                  minimum width=6.8cm, fill=white, blur shadow},
  line/.style  = {-{Latex}, very thick},
  note/.style  = {align=center,font=\footnotesize,inner sep=1pt}
]
\node[block] (stimuli) {%
  \textbf{Synthetic Coffee Stimuli \& Configs}\\
  \texttt{coffee\_scenarios.csv, attribute\_schema.json}\\
  \texttt{kantian\_rules.yml, utilitarian\_weights.yml}\\
  \texttt{explanation\_templates.csv, experiment\_config.yml}
};

\node[block, below=of stimuli] (kb) {%
  \textbf{Ethical Knowledge Base}\\
  Harmonized attribute ranges \& units
};

\node[block, below=of kb] (engines) {%
  \textbf{Dual Reasoning Engines}\\
  Kantian rule engine (deontic violations, severity)\\
  Utilitarian scorer (MCDA over normalized attributes)
};

\node[block, below=of engines] (meta) {%
  \textbf{Meta-Explainer}\\
  Conflict detection \& regret-bounded switching\\
  Contrastive/align\-ment rationales
};

\node[block, below=of meta] (ui) {%
  \textbf{Gamified UI (Streamlit)}\\
  6-round budget allocation, per-round explanations
};

\node[block, below=of ui] (logs) {%
  \textbf{Outputs \& Reproducibility}\\
  \texttt{options\_scored.csv, condition\_summary.csv}\\
  Figures \& Table~1; \texttt{play\_log.csv}
};

\draw[line] (stimuli) -- (kb);
\draw[line] (kb) -- (engines);
\draw[line] (engines) -- (meta);
\draw[line] (meta) -- (ui);
\draw[line] (ui) -- (logs);

\node[note, right=3mm of engines.east] (norm) {min--max\\normalization};
\draw[-{Latex}, thick] (norm.west) -- ++(-6mm,0) |- (engines.east);

\end{tikzpicture}
\caption{One-column architecture: synthetic coffee stimuli and ethical KB feed dual symbolic engines (Kantian + utilitarian); a meta-explainer enforces regret-bounded switches and generates rationales for the gamified UI. All runs are logged for exact reproducibility.}
\label{fig:architecture}
\end{figure}

\section{Results, Discussion \& Outlook}
Across eight synthetic coffee scenarios (Table~\ref{tab:cond_summary}, Fig.~\ref{fig:cond_plots}), four explanation conditions yield the following mean welfare uplift vs.\ a price-only baseline: +0.784 (no explanation), +0.719 (Kantian-only), +0.784 (utilitarian-only), and +0.734 (combined + meta). Violation-free shares are 50\%, 100\%, 50\%, and 75\%, respectively. Mean Kantian-violation severity is 0.875 (no/utilitarian), 0.500 (combined + meta), and 0.000 (Kantian-only). The meta-explainer resolves 25\% of Kantian–utilitarian conflicts by switching from a violating utilitarian-best to a violation-free near-parity option within the regret bound.
These results reveal a tunable trade-off: Kantian secures full deontic compliance at a welfare cost; utilitarian maximizes welfare while tolerating rule breaches; combined + meta preserves near-utilitarian welfare while sharply reducing violations via regret-bounded switches. The dual-logic XAI (deontic rules + weighted MCDA) makes normative tensions visible at decision time and clarifies value–explanation alignment. Next steps include a preregistered human study manipulating the regret bound and explanation salience; testing alignment as a moderator and trust as a mediator; broadening scenarios beyond coffee with verifiable evidence (LCA, certifications, CSR); personalizing thresholds and weights online; adding uncertainty-aware and contrastive rationales; comparing virtue and care ethics; and A/B-testing the UI for comprehension and cognitive load—toward principled, actionable consumer guidance.

\section*{Acknowledgment}
This research was undertaken, in part, thanks to funding from the Canada Research Chairs Program.


\begin{thebibliography}{00}
\bibitem{b1}
T. Miller, ``Explanation in artificial intelligence: Insights from the social sciences,'' \emph{Artificial Intelligence}, vol. 267, pp. 1--38, 2019, doi: 10.1016/j.artint.2018.07.007.

\bibitem{b2}
Y. Rong, T. Leemann, T.-T. Nguyen, L. Fiedler, P. Qian, V. Unhelkar, T. Seidel, G. Kasneci, and E. Kasneci, ``Towards human-centered explainable AI: A survey of user studies for model explanations,'' \emph{IEEE Transactions on Pattern Analysis and Machine Intelligence}, vol. 46, no. 4, pp. 2104--2122, 2024, doi: 10.1109/TPAMI.2023.3331846.
\bibitem{b3}
Z.~Atf and P.~R.~Lewis, ``Is trust correlated with explainability in AI? A meta-analysis,'' \emph{IEEE Transactions on Technology and Society}, 2025.

\bibitem{b4}
Z.~Atf and P.~R.~Lewis, ``Human Centricity in the Relationship Between Explainability and Trust in AI,'' \emph{IEEE Technology and Society Magazine}, vol.~42, no.~4, pp.~66--76, Dec.~2023.

\bibitem{b5}
Z.~Atf, S.~A.~A.~Safavi-Naini, P.~R.~Lewis, A.~Mahjoubfar, N.~Naderi, T.~R.~Savage, and A.~Soroush, ``The challenge of uncertainty quantification of large language models in medicine,'' \emph{arXiv preprint arXiv:2504.05278}, 2025.
\bibitem{b6}
K. Aquino and A. Reed II, ``The self-importance of moral identity,'' \emph{Journal of Personality and Social Psychology}, vol. 83, no. 6, pp. 1423--1440, 2002, doi: 10.1037/0022-3514.83.6.1423.
\bibitem{b7}
Z.~Atf, F.~Taherikia, and K.~Heidarzadeh Hanzaei, ``Designing a Model for Brand Engagement Value Creation through the Integration of Gamification Technology and Explainable Artificial Intelligence (XAI),'' \emph{Journal of Value Creating in Business Management}, vol.~4, no.~3, pp.~340--365, 2024, doi: 10.22034/jvcbm.2024.434941.1292.
\bibitem{b8}
Z.~Atf, F.~Taherikia, and K.~Heidarzadeh Hanzaee, ``Modeling Brand Engagement in Social Media (Based on Sentiment Analysis and Customer Data),'' \emph{International Journal of Innovation Management and Organizational Behavior (IJIMOB)}, vol.~3, no.~2, pp.~208--218, 2023, doi: 10.61838/kman.ijimob.3.2.25.

\bibitem{b9}
J. Hamari, J. Koivisto, and H. Sarsa, ``Does gamification work? A literature review of empirical studies on gamification,'' in \emph{Proc. 47th Hawaii Int. Conf. System Sciences (HICSS)}, 2014, pp. 3025--3034, doi: 10.1109/HICSS.2014.377.

\end{thebibliography}
\end{document}